\documentclass[runningheads]{llncs}
\usepackage{amsmath}
\usepackage{amssymb}
\usepackage{subfigure}
\usepackage{multirow} 
\usepackage{graphicx}
\usepackage[table]{xcolor}
\usepackage{lscape}
\usepackage{microtype}
\usepackage[title]{appendix}

\begin{document}
\title{A Sentence-level Hierarchical BERT Model for Document Classification with Limited Labelled Data\thanks{Supported by organization Teagasc and Insight Centre for Data Analytics.}}
%
%
\author{Jinghui Lu\inst{1}\orcidID{0000-0001-7149-6961} \and
Maeve Henchion\inst{2}\orcidID{0000-0001-7842-1516} \and
Ivan Bacher\inst{3} \and
Brian Mac Namee\inst{1}\orcidID{0000-0003-2518-0274}}
\authorrunning{Lu et al.}
%
\institute{University College Dublin, Dublin, Ireland \email{Jinghui.Lu@ucdconnect.ie, Brian.MacNamee@ucd.ie}\and
Teagasc Agriculture and Food Development Authority, Dublin, Ireland
\email{Maeve.Henchion@teagasc.ie }\\
\and
ADAPT Research Centre, Dublin, Ireland\\
\email{Ivan.Bacher@gmail.com }}
%
\maketitle              
\begin{abstract}
Training deep learning models with limited labelled data is an attractive scenario for many NLP tasks, including document classification. While with the recent emergence of BERT, deep learning language models can achieve reasonably good performance in document classification with few labelled instances, there is a lack of evidence in the utility of applying BERT-like models on long document classification. This work introduces a long-text-specific model --- the Hierarchical BERT Model (HBM) --- that learns sentence-level features of the text and works well in scenarios with limited labelled data. Various evaluation experiments have demonstrated that HBM can achieve higher performance in document classification than the previous state-of-the-art methods with only 50 to 200 labelled instances, especially when documents are long. Also, as an extra benefit of HBM, the salient sentences identified by learned HBM are useful as explanations for labelling documents based on a user study.

\end{abstract}
\section{Introduction}\label{sec:introduction}

In many real-world applications, it is not often that people working in NLP can access sufficiently large labelled data to train deep learning models. With the recent advance in Bidirectional Encoder Representations from Transformers (BERT) and its variants \cite{devlin2018bert,liu2019roberta}, language models can extract extensive general language features from large corpora like never before and transfer learned knowledge to the target domain where labelled data is limited. Previous work has demonstrated that the pre-training and fine-tuning paradigm based on BERT outperforms the traditional NLP approaches in document classification where people have limited number of labelled examples \cite{usherwood2019low}. However, we argue that when texts are long, even BERT can not generate proper results with a small training data due to the text length limitation of BERT \cite{ding2020cogltx}.

Inspired by \emph{Hierarchical Attention Network} (HAN) \cite{cho2014learning} that extracts sentence-level information for document classification, in this work, we attempt to further improve the performance of BERT-based models in document classification under low labelled data context by considering the sentence structure information of texts. Specifically, following the intrinsic attention mechanism of the BERT model, which is good at capturing the relationship between words, we propose a \emph{Hierarchical BERT Model} (HBM) which is focused on the sentence-level structure of a document. In the same way that BERT captures the connections between words, the HBM is designed to capture the connections between sentences. We believe that this approach can overcome the loss of information regarding document structure that occurs when simple averaging of word vectors (such as word2vec, BERT-based language models) is used to form the document representations. We expect that the proposed model can increase classification accuracy especially for long documents. Furthermore, due to the sentence attention mechanism in HBM, not all sentences are equally weighted when constructing the document representation. In HBM, information from sentences that receive higher attention scores is more likely to be aggregated to form the representation of the document. We propose that sentences contributing more to the representation of the document are important sentences, and that these sentences can be used as explanatory sentences to make it easier for the human to make a decision about a document when performing the labelling task. Hence, in this paper, we raise the following research questions:

\begin{description}

\item{{\fontseries{b}\selectfont RQ1:}} \textit{Is the proposed sentence-level BERT-based model, the Hierarchical BERT Model (HBM), able to achieve high classification performance in scenarios with small amounts of labelled data for training, especially when documents are long?}

\item{{\fontseries{b}\selectfont RQ2:}} \textit{Can highlighting important sentences identified by the HBM help the annotator label documents faster?}

\end{description}

%
%
%
%


%
%
%

The remainder of the paper is organised as follows: Section \ref{sec:hierarchical_related_work} summarises the related work; Section \ref{sec:hierarchical_methods} describes the proposed HBM method in detail; Section \ref{sec:hierarchical_exp2} describes an evaluation experiment comparing the proposed HBM method with state-of-the-art methods under the low labelled data scenario; Section \ref{sec:hierarchical_exp3} introduces the inference of important sentences by HBM; Section \ref{sec:hierarchical_exp4} demonstrates the effectivenss of the HBM method in selecting important sentences to show to human by a user study; and Section \ref{sec:hierarchical_summary} draws conclusions.

\section{Related Work}\label{sec:hierarchical_related_work}

This work seeks to build on the BERT language models to further explore the effectiveness of taking sentence-level information into account for document classification with limited labelled data. Though BERT and its variants have achieved impressive results in many NLP tasks, including document classification in larger datasets \cite{devlin2018bert,liu2019roberta}, few studies are focused on the performance of BERT in low labelled data scenario over other existing NLP paradigms. As suggested in \cite{goodfellow2016deep}, supervised deep learning models will generally achieve acceptable performance at about 5,000 examples per class, and therefore, deep learning models may not be suitable for scenarios with low labelled examples. Interestingly, Usherwood and Smit \cite{usherwood2019low} demonstrated that when one has only 100-1,000 labelled examples per class, BERT is the best performing approach for document classification as compared to other traditional machine learning approaches, due to the utility of pre-training and fine-tuning paradigm. However, their work is an empirical comparison of BERT and other NLP approaches without any modification in BERT. Also, BERT is incapable of processing long texts \cite{ding2020cogltx}, the datasets used in \cite{usherwood2019low} are Amazon reviews and Twitter that consist of short texts, lacking evidence of the utility of BERT-based models for long text classification with limited labelled data. Hence, our work proposes using a sentence-level language model to alleviate the long-texts problem suffered by BERT.

Yang et al. \cite{yang2016hierarchical} were the first to propose the sentence-level model \emph{Hierarchical Attention Network} (HAN) for document classification, and they adopted the Bi-GRU with attention as their architecture. There are also some concurrent works investigating the effectiveness of sentence-level model based on BERT. For example, Xu et al. \cite{xu2019many} used a sentence-level BERT to a query matching task in Chinese (judge whether the two sentences convey the same meaning), Yang et al. \cite{yang2020html} proposed a sentence-level BERT integrated with audio features to forecast the price of a financial asset over a certain period, and Zhang et al. \cite{zhang2019hibert} designed HIerachical Bidirectional Encoder Representations from Transformers (HIBERT) for document summarisation. Though all used sentence-level information, these works vary from the concrete design, pre-training/fine-tuning paradigm, the datasets used, the task performed and the domain. Moreover, these studies assume an abundance of labelled data instead of focusing on the limited labelled data scenarios. 


The contributions of this paper are: (1) we propose a sentence-level BERT-based model, the \emph{Hierarchical BERT Model}, that captures the sentence structure information in a document and an evaluation experiment that demonstrates that HBM is effective in scenarios with small amounts of labelled data; (2) We demonstrate that the salient sentences identified by the HBM are useful as explanations for document classification, and can help annotators label documents faster based on a user study.

\section{The Hierarchical BERT Model}\label{sec:hierarchical_methods}

\begin{figure*}[hbt!]
\includegraphics[width=1.0\textwidth]{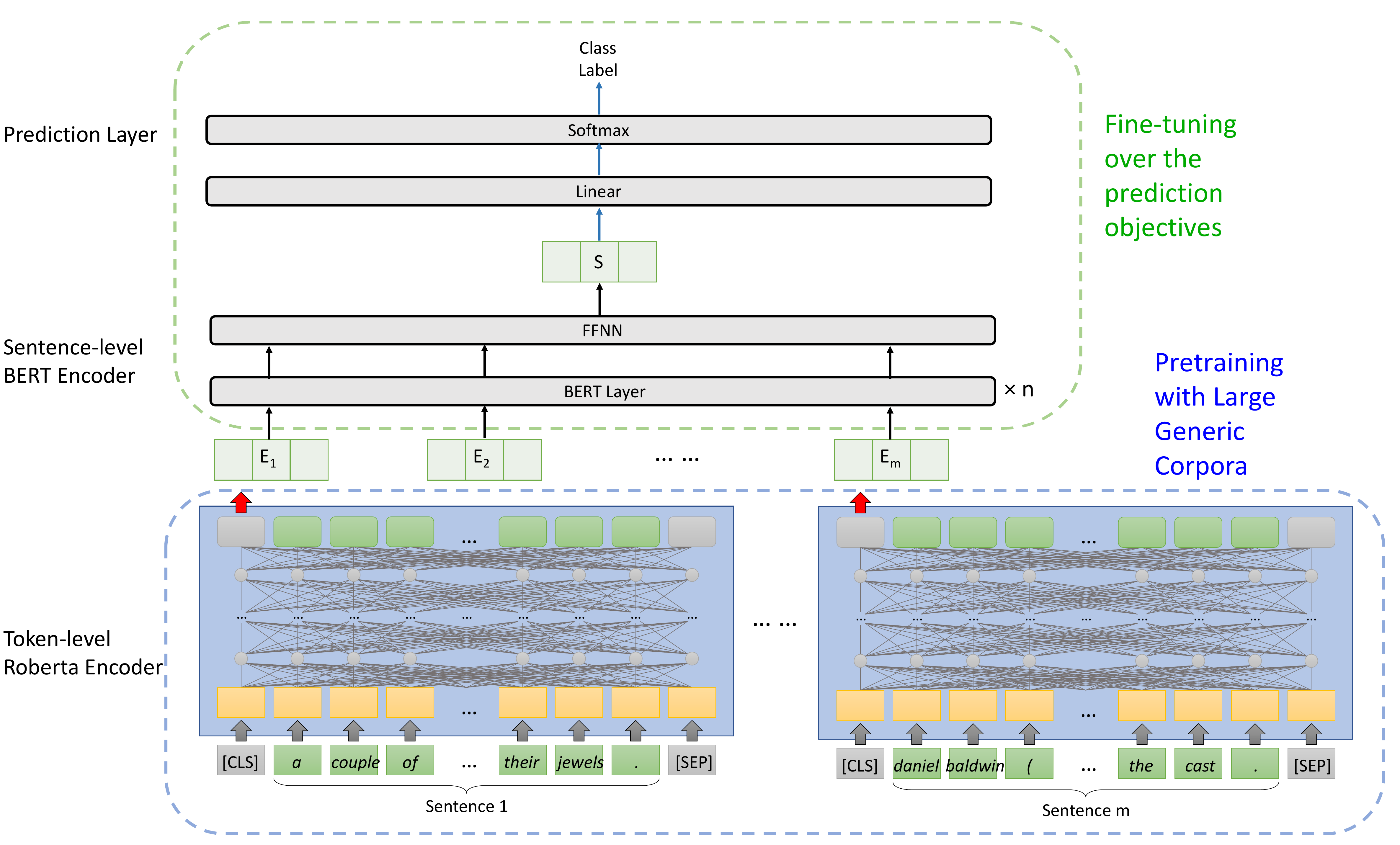}
 \caption{The Hierarchical BERT Model for document classification.}
\label{fig:hierarchical_model}
\end{figure*}

Figure \ref{fig:hierarchical_model} summarises the proposed sentence-level Hierarchical BERT Model for document classification which consists of 3 components: (1) the token-level Roberta encoder; (2) the sentence-level BERT encoder; and (3) the prediction layer. To make predictions using this model, raw text is first fed into a token-level Roberta encoder to extract text features and form the vector representation for each sentence. Subsequently, the sentence vectors generated by the token-level Roberta encoder are used as input for the sentence-level BERT encoder. Finally, the intermediate representation generated by the sentence-level BERT encoder is input into the prediction layer that predicts the document class. The detailed implementations of each component are described in the following sections.

\vspace{-5mm}

\subsection{Token-level Roberta Encoder}\label{subsec:hierarchical_token}

Using the same architecture of BERT but optimising the key hyper-parameters and pre-training with larger datasets, Roberta \cite{liu2019roberta} is recognised as a state-of-the-art approach across many NLP tasks and adopted as the word-level foundation of HBM. Briefly, the multi-head self-attention mechanism in Roberta can effectively capture the semantic and syntactic information between words in a document. In this work, we leverage a pre-trained Roberta model to extract raw text features which act as the input to a sentence-level BERT encoder. The output of this component can be represented as follows:

\begin{equation}\mathcal{D}= (E_{1}, E_{2}, \ldots, E_{m} )\end{equation}

\noindent Here  $E_{i}$ denotes the $i^{th}$ sentence vector (of dimension $d_{e}$ by averaging token embeddings contained in each sentence) of document $\mathcal{D} \in \mathbb{R}^{m \times d_{e}}$ generated by the token-level Roberta encoder, where $m$ is the maximum number of sentences set by the user (padding with zeros if the number of sentences is less than $m$).

\subsection{Sentence-level BERT Encoder}\label{subsec:hierarchical_sentence}

\begin{figure*}
\centering
\includegraphics[width=.7\textwidth]{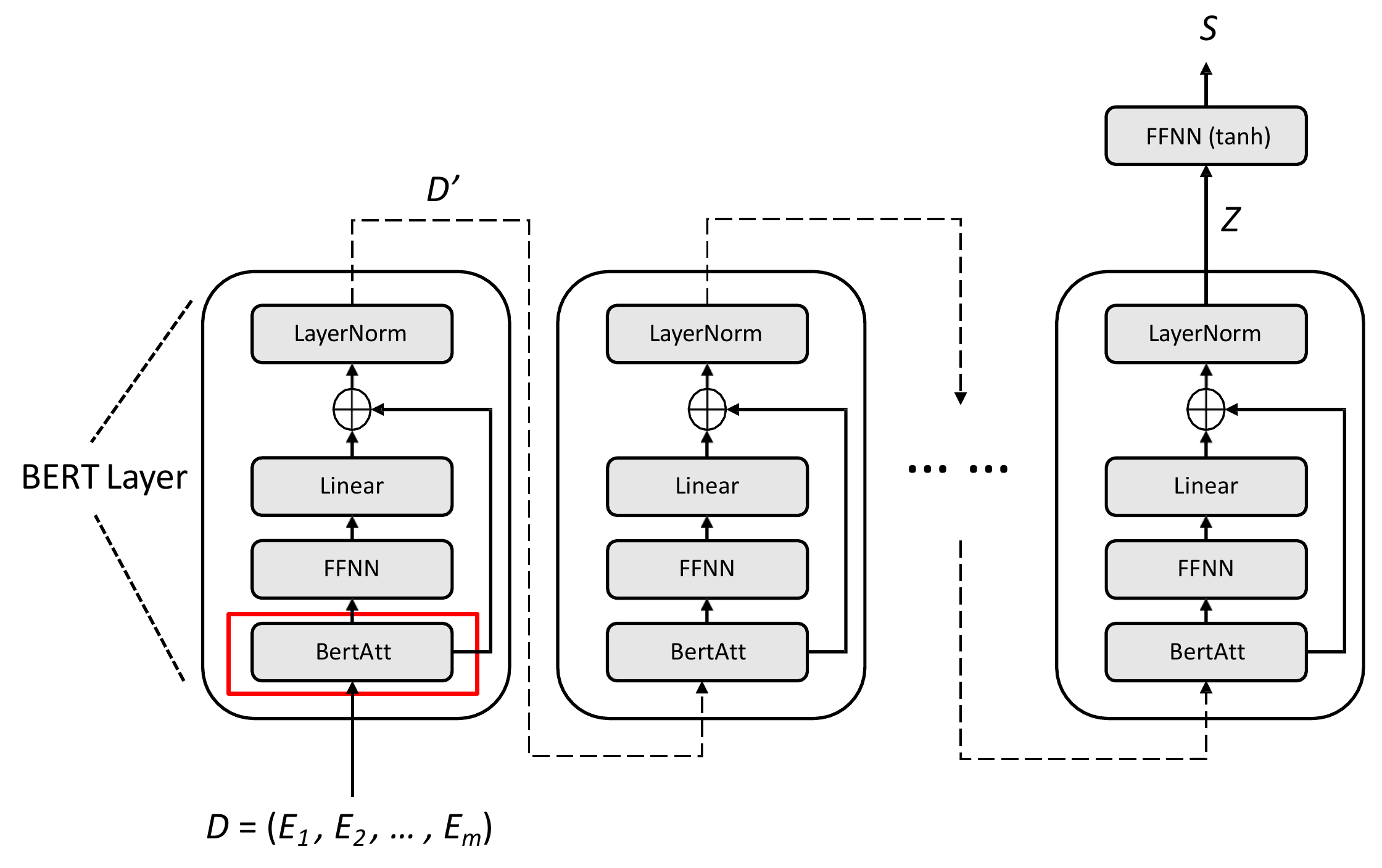}
 \caption{The BERT layers in the sentence-level BERT encoder.}
\label{fig:hierarchical_bert_layer}
\end{figure*}

The sentence-level Hierarchical BERT encoder extracts the sentence structure information for generating an intermediate document representation, $\mathcal{S}$, based on the sentence vectors, $\mathcal{D}$, output by the token-level Roberta encoder. As shown in Figure \ref{fig:hierarchical_bert_layer}, the sentence-level BERT encoder consists of several identical BERT layers and one feedforward neural network with a tanh activation. Initially, sentence vectors $\mathcal{D}$ from the token-level Roberta encoder are input together into a BERT layer, more specifically, the BertAtt layer where the multi-head self-attention mechanism is applied (denoted by a red rectangle in Figure \ref{fig:hierarchical_bert_layer}). 

\vspace{-5mm}

\vspace{-5mm}

The output of the BertAtt layer is calculated as follows:

\begin{subequations}
\begin{align}
\text{ BertAtt} (\mathcal{D}   )=\text{ LayerNorm }(\mathcal{D}+\text{MultiHead}(\mathcal{D}))\label{eq:bertAttention} \\
\text {MultiHead} (\mathcal{D} ) =\text {Concat} (\text {head}_{1},\text{head}_{2},\ldots, \text {head}_{h} ) \times W^{O}
\end{align}
\end{subequations}

where $h$ is the number of heads in the BertAtt layer; $\text{LayerNorm}$ is layer normalization; $W^{O} \in \mathbb{R}^{hd_{e} \times d_{e}}$ is the weight matrix for dimension transformation; and $\text{head}_{i}$ is the attention of the $i^{th}$ head. Each  $\text{head}_{i}$ value is calculated as follows:

\begin{subequations}
\begin{align}
\text {head}_{i}&= \text {Attention}(Q_{i},K_{i},V_{i})=\text{Softmax}(Q_{i}\times K_{i}^{\top}/\sqrt{d_{e}})\times V_{i} \\
Q_{i}&=\mathcal{D}\times W_{i}^{Q} \qquad K_{i}=\mathcal{D}\times W_{i}^{K} \qquad  V_{i}=\mathcal{D}\times W_{i}^{V}
\end{align}
\label{eq:attention_score}
\end{subequations}

where $W_{i}^{Q}, W_{i}^{K}, W_{i}^{V} \in \mathbb{R}^{d_{e} \times d_{e}}$ are weight matrices for the $i^{th}$ head; and, $ \text{Softmax}(Q_{i}\times K_{i}^{\top}/\sqrt{d_{e}})$ is an $m\times m$ matrix in which the entry at the $a^{th}$ row and $j^{th}$ column denotes the attention weight that the $a^{th}$ sentence pays to the $j^{th}$ sentence. Here, $V_{i}$ contains information from sentences to pass through the following layers while the attention weights matrix, $ \text{Softmax}(Q_{i}\times K_{i}^{\top}/\sqrt{d_{e}})$, acts as a gate to control how much information can be passed through (i.e. after the multiplication, it is hard for sentences with low attention score to further propagate their information).

Then, the output of BertAtt is passed through a standard feedforward neural network with the residual mechanism and layer normalisation which is computed as:

\begin{equation}\begin{aligned}
\mathcal{D'} &=\text{LayerNorm} ( \text {BertAtt} (\mathcal{D} )+\text{Relu}( \text {BertAtt} (\mathcal{D} )\times W^{r})\times W^{S} )
\label{eq:hierarchical_B}
\end{aligned}\end{equation}

\noindent where $ \text {BertAtt} (\mathcal{D} )\in \mathbb{R}^{m \times d_{e}}$ is the output of the BertAtt layer calculated by Equation \ref{eq:bertAttention}; $W^{r} \in \mathbb{R}^{d_{e} \times nd_{e}}$ is a weight matrix that transforms the dimension of $\text {BertAtt} (\mathcal{D} )$ from $d_{e}$ to $nd_{e}$ (we set $n$ to 4 in our work); and $W^{S} \in \mathbb{R}^{nd_{e} \times d_{e}}$ is the weight matrix with dropout to transform the dimension back to $d_{e}$. $\mathcal{D'}$ then goes into multiple identical BERT layers to form the matrix $\mathcal{Z}$ which is used for computing the intermediate document representation $\mathcal{S}$.

The intermediate document representation $\mathcal{S}\in \mathbb{R}^{1 \times d_{e}}$ output by the sentence-level BERT encoder can be computed as (see Figure \ref{fig:hierarchical_bert_layer}):


\begin{equation}
\mathcal{S}=\text{Tanh} (\text{Avg}(\mathcal{Z}) \times W^{t} )
\end{equation}

\noindent where $W^{t}\in \mathbb{R}^{d_{e} \times d_{e}}$ is a weight matrix for linear transformation, $\mathcal{Z}\in \mathbb{R}^{m \times d_{e}}$ is the output of the multiple BERT layers, $\text{Avg}$ denotes the mean pooling layer (i.e. $\text{Avg}(\mathcal{Z})\in \mathbb{R}^{1 \times d_{e}}$).

Finally, the output of the sentence-level BERT Encoder, $\mathcal{S}$, is passed to the prediction layer.

\subsection{Prediction Layer}\label{subsec:hierarchical_prediction}

The prediction layer is appended to the sentence-level BERT encoder for final prediction as shown in Figure \ref{fig:hierarchical_model}. The output, $\mathcal{S}$, from the sentence-level BERT encoder is used to compute the raw score for each class, the process can be shown as:

\begin{equation}
[t_{0},t_{1}, ..., t_{y}] = \mathcal{S}\times W \label{eq:hierarchical_t}
\end{equation}

\noindent where $W \in \mathbb{R}^{d_{e} \times y}$ is a weight matrix and $t_{0}, t_{1}, ..., t_{y}$ represents the raw score of the document for the $0^{th}$ to $y^{th}$ class respectively which is fed into a softmax head later for computing loss.

\subsection{Training a Hierarchical BERT Model (HBM)}\label{subsec:hierarchical_training}

It is common for BERT-like models to exploit the pre-training + fine-tuning procedure to train the model \cite{devlin2018bert,liu2019roberta}. More specifically, in pre-training, the model is trained via several self-supervised methods with various, large, unlabelled datasets. Then in fine-tuning, a task-specific head is added to the model to further update all parameters of the model by supervised learning over small local datasets.

However, the pre-training and fine-tuning is a little different in the proposed HBM approach. Since the Roberta encoder used in HBM is pre-trained with large generic corpora, we assume that the Roberta encoder can encode the token-level information well. Hence, we do not further adjust the pre-trained Roberta model during fine-tuning. In other words, in fine-tuning, we freeze the weights of the Roberta encoder while weights in the sentence-level BERT encoder, as well as in the prediction layer, are randomly initialised and updated based on our prediction objectives. This is illustrated in Figure \ref{fig:hierarchical_model} through the dashed green and blue rectangles.  

The objective function optimised during training is a weighted cross entropy:

\begin{equation}\text{loss}(t,y)=weight[y]( -\log (\exp (t_{y})/\sum_{j}^{z} \exp (t_{j})))\end{equation}

where $y\in [0,1, ..., z]$ is the index of the ground truth class,  $t_{y}$ denotes the raw, unnormalised scores for the $y^{th}$ class computed by Equation \ref{eq:hierarchical_t} and $weight[y]$ implies the factor given to the $y^{th}$ class proportional to the class ratio. 

\section{Experiment 1: Evaluating the Effectiveness of HBM}\label{sec:hierarchical_exp2}

This section describes the design of an experiment performed to evaluate the performance of HBM for document classification, especially when training datasets are small, by comparing it to other state-of-the-art methods. This addresses the first research question, {\fontseries{b}\selectfont RQ1}. The following subsections describe the datasets used in the experiments, the set up of several baselines as well as HBM, and the experimental results.

\subsection{Datasets}\label{subsec:hierarchical_exp2_datasets}

We evaluate the effectiveness of our model on six fully-labelled binary document classification datasets that primarily consist of long documents (from news feeds, blogs, etc.), we also include datasets that consist of short documents to investigate the performance of HBM on short texts:

\begin{itemize}
\item {\fontseries{b}\selectfont Movie Reviews:} 1,000 \textit{positive} sentiment and 1,000 \textit{negative} sentiment processed movie reviews collected from \textit{rottentomatoes.com} \cite{pang2004sentimental}. The average number of sentences per document is 33.12 and the maximum number of sentences is 114.
\item {\fontseries{b}\selectfont Multi-domain Customer Review:} 3,999 \textit{negative} sentiment and 4,000 \textit{positive} sentiment processed customer reviews collected from \textit{Amazon} from the books, DVDs, electronics and kitchen \& housewares sections \cite{blitzer2007biographies}. The average number of sentences per document is 3.78, these are not long documents, and the maximum number of sentences is 181.
\item {\fontseries{b}\selectfont Blog Author Gender:} 1,678 \textit{male} blogs and 1,548 \textit{female} blogs \cite{mukherjee2010improving}. The average number of sentences per document is 22.83 and the maximum number of sentences is 511.
\item {\fontseries{b}\selectfont Guardian 2013:} 843 \textit{politics} news and 843 \textit{business} news items collected from the Guardian across 2013 \cite{belford2018stability}. The average number of sentences per document is 27.88 and the maximum number of sentences is 724.
\item {\fontseries{b}\selectfont Reuters:} 2,369 \textit{acq} news items and 2,369 \textit{earn} news items. The average number of sentences per document is 7.40, these are not long documents, and the maximum number of sentences is 95.
\item {\fontseries{b}\selectfont 20 Newsgroups:} 578 \textit{comp.sys.mac.hardware} newsgroup documents and 590 \textit{comp.sys.pc.hardware} newsgroup documents. The average number of sentences per document is 10.61 and the maximum number of sentences is 367.
\end{itemize}

\subsection{Baselines and Setup}\label{subsec:hierarchical_exp2_baseline}
Partially following \cite{usherwood2019low}, we compare HBM with several baseline methods. We selected an SVM model using a document representation based on FastText, an SVM model using text representations based on a pre-trained Roberta model and a fine-tuned Roberta model as baselines due to their strong performance. We also used another sentence-level model, Hierarchical Attention Network (HAN) \cite{yang2016hierarchical}, as a baseline. 

\noindent{\fontseries{b}\selectfont FastText + SVM:} We use 300-dimensional word vectors constructed by a FastText language model pre-trained with the Wikipedia corpus \cite{joulin2016fasttext}. Averaged word embeddings are used as the representation of the document. For preprocessing, all text is converted to lowercase and we remove all punctuation and stop words. SVM is used as the classifier. We tune the hyper-parameters of the SVM classifier using a grid-search based on 5-fold cross-validation performed on the training set, after that, we re-train the classifier with optimised hyper-parameters. This hyper-parameter tuning method is applied in Roberta + SVM as well.

\noindent{\fontseries{b}\selectfont Roberta + SVM:} We use 768-dimensional word vectors generated by a pre-trained Roberta language model \cite{liu2019roberta}. We do not fine-tune the pre-trained language model and use the averaged word vectors as the representation of the document. Since all BERT-based models are configured to take as input a maximum of 512 tokens, we divided the long documents with $W$ words into $k = W/511$ fractions, which is then fed into the model to infer the representation of each fraction (each fraction has a ``[CLS]'' token in front of 511 tokens, so, 512 tokens in total). Based on the approach of \cite{sun2019fine}, the vector of each fraction is the average embeddings of words in that fraction, and the representation of the whole text sequence is the mean of all $k$ fraction vectors. For preprocessing, the only operation performed is to convert all tokens to lowercase. SVM is used as the classifier.

\noindent{\fontseries{b}\selectfont Fine-tuned Roberta:} For the document classification task, fine-tuning Roberta means adding a softmax layer on top of the Roberta encoder output and fine-tuning all parameters in the model.  In this experiment, we fine-tune the same 768-dimensional pre-trained Roberta model with a small training set. The settings of all hyper-parameters follow \cite{liu2019roberta}. we set the learning rate to $1\times10^{-4}$ and the batch size to 4, and use the Adam optimizer \cite{kingma2014adam} with $\epsilon$ equals to $1\times10^{-8}$ through hyperparameter tuning. However, since we assume that the amount of labelled data available for training is small, we do not have the luxury of a hold out validation set to use to implement early stopping during model fine tuning. Instead, after training for 15 epochs we roll back to the model with the lowest loss based on the training dataset. This rollback strategy is also applied to HAN and HBM due to the limited number of instances in training sets. For preprocessing, the only operation performed is to convert all tokens to lowercase.

\noindent{\fontseries{b}\selectfont Hierarchical Attention Network:} Following \cite{yang2016hierarchical}, we apply two levels of Bi-GRU with attention mechanism for document classification. All words are first converted to word vectors using GloVe \cite{pennington2014glove} (300 dimension version pre-trained using the wiki gigaword corpus) and fed into a word-level Bi-GRU with attention mechanism to form sentence vectors. After that, a sentence vector along with its context sentence vectors are input into sentence-level Bi-GRU with attention mechanism to form the document representation which is then passed to a softmax layer for final prediction. For preprocessing, the only operation performed is to convert all tokens to lowercase, and separate documents into sentences.\footnote{We apply Python NLTK sent\_tokenize function to split documents into sentences.}

\noindent{\fontseries{b}\selectfont Hierarchical BERT Model:} For HBM, we set the number of BERT layers to 4, and the maximum number of sentences to 114, 64, 128, 128, 100, and 64 for the \emph{Movie Review, Multi-domain Customer Review, Blog Author Gender, Guardian 2013, Reuters} and \emph{20 Newsgroups} datasets respectively, these values are based on the length of documents in these datasets. After some preliminary experiments, we set the attention head to 1, the learning rate to $2\times10^{-5}$, dropout probability to 0.01, used 50 epochs, set the batch size to 4 and used the Adam optimizer with $\epsilon$ equals to $1\times10^{-8}$. The only text preprocessing operation performed is to convert all tokens to lowercase and split documents into sentences. 

To simulate the scenarios of limited training data, we randomly subsample a small set of data from the fully labelled dataset as the training data. For each dataset, we subsample 50, 100, 150 and 200 instances respectively and we always use the full dataset minus 200 instances as the test set. For each experiment, the training is repeated 10 times with training datasets subsampled by 10 different random seeds and we report the averaged result of 10 repetitions. We use area under the ROC curve (AUC) to measure the classification performance.

\begin{table}[ht]
\centering
\caption{Performance measured using AUC score of different methods that are trained with limited labelled data across six datasets. }
\label{tab:hierarchical_exp2_auc}
\resizebox{.8\textwidth}{!}{%
\begin{tabular}{l|llllc}
\hline
{\fontseries{b}\selectfont Dataset} & \multicolumn{5}{c}{{\fontseries{b}\selectfont Movie Review}} \\
{\fontseries{b}\selectfont \#Instances} & {\fontseries{b}\selectfont n = 50} & {\fontseries{b}\selectfont n = 100} & {\fontseries{b}\selectfont n = 150} & {\fontseries{b}\selectfont n = 200} & \multicolumn{1}{l}{{\fontseries{b}\selectfont Avg Rank}} \\ \hline
{\fontseries{b}\selectfont FastText} & 0.6653$\pm$0.171 (5) & 0.7942$\pm$0.020 (4) & 0.8040$\pm$0.018 (4) & 0.8260$\pm$0.010 (4) & 4.25 \\
{\fontseries{b}\selectfont Roberta} & 0.8743$\pm$0.032 (3) & 0.9132$\pm$0.025 (3) & 0.9397$\pm$0.010 (3) & 0.9449$\pm$0.008 (3) & 3 \\
{\fontseries{b}\selectfont Fine-tuned Roberta} & 0.8878$\pm$0.018 (2) & 0.9298$\pm$0.013 (2) & 0.9451$\pm$0.007 (2) & 0.9536$\pm$0.005 (2) & 2 \\
{\fontseries{b}\selectfont HAN} & 0.7013$\pm$0.096 (4) & 0.7789$\pm$0.014 (5) & 0.7504$\pm$0.082 (5) & 0.8128$\pm$0.011 (5) & 4.75 \\
{\fontseries{b}\selectfont HBM} & {\fontseries{b}\selectfont 0.9139$\pm$0.020 (1)} & {\fontseries{b}\selectfont 0.9420$\pm$0.011 (1)} & {\fontseries{b}\selectfont 0.9572$\pm$0.007 (1)} & {\fontseries{b}\selectfont 0.9638$\pm$0.006 (1)} & {\fontseries{b}\selectfont 1} \\ \hline
{\fontseries{b}\selectfont Dataset} & \multicolumn{5}{c}{{\fontseries{b}\selectfont Blog Author Gender}} \\
{\fontseries{b}\selectfont \#Instances} & {\fontseries{b}\selectfont n = 50} & {\fontseries{b}\selectfont n = 100} & {\fontseries{b}\selectfont n = 150} & {\fontseries{b}\selectfont n = 200} & \multicolumn{1}{l}{{\fontseries{b}\selectfont Avg Rank}} \\ \hline
{\fontseries{b}\selectfont FastText} & 0.6400$\pm$0.043 (4) & 0.6669$\pm$0.021 (5) & 0.6899$\pm$0.009 (4) & 0.6861$\pm$0.023 (5) & 4.5 \\
{\fontseries{b}\selectfont Roberta} & 0.5538$\pm$0.153 (5) & 0.6783$\pm$0.025 (3) & 0.7058$\pm$0.017 (3) & 0.7213$\pm$0.012 (3) & 3.5 \\
{\fontseries{b}\selectfont Fine-tuned Roberta} & 0.6462$\pm$0.036 (2) & 0.6892$\pm$0.021 (2) & 0.7177$\pm$0.019 (2) & 0.7295$\pm$0.024 (2) & 2 \\
{\fontseries{b}\selectfont HAN} & 0.6402$\pm$0.017 (3) & 0.6670$\pm$0.010 (4) & 0.6845$\pm$0.008 (5) & 0.6876$\pm$0.014 (4) & 4 \\
{\fontseries{b}\selectfont HBM} & {\fontseries{b}\selectfont 0.6820$\pm$0.025 (1)} & {\fontseries{b}\selectfont 0.7150$\pm$0.031 (1)} & {\fontseries{b}\selectfont 0.7371$\pm$0.007 (1)} & {\fontseries{b}\selectfont 0.7488$\pm$0.013 (1)} & {\fontseries{b}\selectfont 1} \\ \hline
{\fontseries{b}\selectfont Dataset} & \multicolumn{5}{c}{{\fontseries{b}\selectfont Reuters}} \\
{\fontseries{b}\selectfont \#Instances} & {\fontseries{b}\selectfont n = 50} & {\fontseries{b}\selectfont n = 100} & {\fontseries{b}\selectfont n = 150} & {\fontseries{b}\selectfont n = 200} & \multicolumn{1}{l}{{\fontseries{b}\selectfont Avg Rank}} \\ \hline
{\fontseries{b}\selectfont FastText} & 0.9757$\pm$0.010 (4) & 0.9795$\pm$0.005 (5) & 0.9851$\pm$0.004 (5) & 0.9862$\pm$0.003 (5) & 4.75 \\
{\fontseries{b}\selectfont Roberta} & 0.9838$\pm$0.007 (2) & 0.9890$\pm$0.003 (3) & 0.9931$\pm$0.001 (3) & 0.9930$\pm$0.001 (3) & 2.75 \\
{\fontseries{b}\selectfont Fine-tuned Roberta} & {\fontseries{b}\selectfont 0.9885$\pm$0.005 (1)} & {\fontseries{b}\selectfont 0.9933$\pm$0.002 (1)} & 0.9953$\pm$0.001 (2) & 0.9955$\pm$0.001 (2) & {\fontseries{b}\selectfont 1.5} \\
{\fontseries{b}\selectfont HAN} & 0.9270$\pm$0.038 (5) & 0.9804$\pm$0.005 (4) & 0.9865$\pm$0.003 (4) & 0.9897$\pm$0.002 (4) & 4.25 \\
{\fontseries{b}\selectfont HBM} & 0.9825$\pm$0.008 (3) & 0.9917$\pm$0.004 (2) & {\fontseries{b}\selectfont 0.9980$\pm$0.003 (1)} & {\fontseries{b}\selectfont 0.9990$\pm$0.001 (1)} & 1.75 \\ \hline
{\fontseries{b}\selectfont Dataset} & \multicolumn{5}{c}{{\fontseries{b}\selectfont Multi-domain Customer Review}} \\
{\fontseries{b}\selectfont \#Instances} & {\fontseries{b}\selectfont n = 50} & {\fontseries{b}\selectfont n = 100} & {\fontseries{b}\selectfont n = 150} & {\fontseries{b}\selectfont n = 200} & \multicolumn{1}{l}{{\fontseries{b}\selectfont Avg Rank}} \\ \hline
{\fontseries{b}\selectfont FastText} & 0.6694$\pm$0.047 (4) & 0.6927$\pm$0.030 (4) & 0.7226$\pm$0.030 (5) & 0.7471$\pm$0.018 (5) & 4.5 \\
{\fontseries{b}\selectfont Roberta} & 0.8317$\pm$0.020 (2) & 0.8558$\pm$0.036 (2) & 0.8976$\pm$0.019 (2) & 0.9190$\pm$0.006 (2) & 2 \\
{\fontseries{b}\selectfont Fine-tuned Roberta} & {\fontseries{b}\selectfont 0.9110$\pm$0.036 (1)} & {\fontseries{b}\selectfont 0.9437$\pm$0.007 (1)} & {\fontseries{b}\selectfont 0.9534$\pm$0.003 (1)} & {\fontseries{b}\selectfont 0.9565$\pm$0.004 (1)} & {\fontseries{b}\selectfont 1} \\
{\fontseries{b}\selectfont HAN} & 0.6497$\pm$0.021 (5) & 0.6907$\pm$0.011 (5) & 0.7312$\pm$0.014 (4) & 0.7739$\pm$0.034 (4) & 4.5 \\
{\fontseries{b}\selectfont HBM} & 0.7669$\pm$0.024 (3) & 0.8342$\pm$0.014 (3) & 0.8615$\pm$0.010 (3) & 0.8913$\pm$0.004 (3) & 3 \\ \hline
{\fontseries{b}\selectfont Dataset} & \multicolumn{5}{c}{{\fontseries{b}\selectfont Gurdian 2013}} \\
{\fontseries{b}\selectfont \#Instances} & {\fontseries{b}\selectfont n = 50} & {\fontseries{b}\selectfont n = 100} & {\fontseries{b}\selectfont n = 150} & {\fontseries{b}\selectfont n = 200} & \multicolumn{1}{l}{{\fontseries{b}\selectfont Avg Rank}} \\ \hline
{\fontseries{b}\selectfont FastText} & 0.9720$\pm$0.003 (3) & 0.9789$\pm$0.005 (5) & 0.9794$\pm$0.004 (5) & 0.9789$\pm$0.005 (5) & 4.5 \\
{\fontseries{b}\selectfont Roberta} & 0.9694$\pm$0.011 (4) & 0.9814$\pm$0.003 (3) & 0.9860$\pm$0.003 (2) & 0.9852$\pm$0.002 (3) & 3 \\
{\fontseries{b}\selectfont Fine-tuned Roberta} & 0.9727$\pm$0.010 (2) & 0.9848$\pm$0.002 (2) & 0.9854$\pm$0.003 (3) & 0.9864$\pm$0.002 (2) & 2.25 \\
{\fontseries{b}\selectfont HAN} & 0.9684$\pm$0.005 (5) & 0.9794$\pm$0.001 (4) & 0.9850$\pm$0.002 (4) & 0.9849$\pm$0.002 (4) & 4.25 \\
{\fontseries{b}\selectfont HBM} & {\fontseries{b}\selectfont 0.9740$\pm$0.013 (1)} & {\fontseries{b}\selectfont 0.9862$\pm$0.007 (1)} & {\fontseries{b}\selectfont 0.9904$\pm$0.003 (1)} & {\fontseries{b}\selectfont 0.9925$\pm$0.001 (1)} & {\fontseries{b}\selectfont 1} \\ \hline
{\fontseries{b}\selectfont Dataset} & \multicolumn{5}{c}{{\fontseries{b}\selectfont 20Newsgroups}} \\
{\fontseries{b}\selectfont \#Instances} & {\fontseries{b}\selectfont n = 50} & {\fontseries{b}\selectfont n = 100} & {\fontseries{b}\selectfont n = 150} & {\fontseries{b}\selectfont n = 200} & \multicolumn{1}{l}{{\fontseries{b}\selectfont Avg Rank}} \\ \hline
{\fontseries{b}\selectfont FastText} & {\fontseries{b}\selectfont 0.7052$\pm$0.030 (1)} & 0.7516$\pm$0.033 (3) & 0.7827$\pm$0.012 (3) & 0.8094$\pm$0.013 (3) & 2.5 \\
{\fontseries{b}\selectfont Roberta} & 0.5969$\pm$0.100 (5) & 0.6988$\pm$0.024 (5) & 0.7117$\pm$0.030 (5) & 0.7436$\pm$0.011 (4) & 4.75 \\
{\fontseries{b}\selectfont Fine-tuned Roberta} & 0.6098$\pm$0.051 (4) & 0.7555$\pm$0.041 (2) & 0.8576$\pm$0.060 (2) & 0.8838$\pm$0.045 (2) & 2.5 \\
{\fontseries{b}\selectfont HAN} & 0.6296$\pm$0.030 (3) & 0.7142$\pm$0.005 (4) & 0.7295$\pm$0.012 (4) & 0.7417$\pm$0.029 (5) & 4 \\
{\fontseries{b}\selectfont HBM} & 0.6883$\pm$0.094 (2) & {\fontseries{b}\selectfont 0.8168$\pm$0.024 (1)} & {\fontseries{b}\selectfont 0.8579$\pm$0.021 (1)} & {\fontseries{b}\selectfont 0.9158$\pm$0.018 (1)} & {\fontseries{b}\selectfont 1.25} \\ \hline
\end{tabular}%
}
\end{table}

\vspace{-5mm}

\subsection{Results and Analysis}\label{subsec:hierarchical_exp2_result}

Table \ref{tab:hierarchical_exp2_auc} summarises the results of the evaluation experiments performed. In Table \ref{tab:hierarchical_exp2_auc}, the leftmost column denotes the modelling approach, the topmost row denotes the datasets used in the experiment, and $n$ is the number of instances in the training set. The numbers in brackets stand for the ranking of each method when compared to the performance of other approaches for a dataset with a specific number of labelled instances. The Avg Rank column denotes the average ranking of each model for a specific dataset. The best performance in each column is highlighted in bold. It is clear from Table \ref{tab:hierarchical_exp2_auc} that the performance of each method improves as the number of instances in the training set increases. The performance of the proposed HBM method consistently surpasses the performance of the other state-of-the-art methods in the majority of experiments. This clearly demonstrates the effectiveness of the HBM approach under the scenario of limited labelled data. 

The performance of HBM is not good for the \emph{Multi-domain Customer Review} and \emph{Reuters} datasets (even lower than that of the approach based on the pre-trained Roberta model for the former). We hypothesise that this is caused by the low number of sentences per document in these datasets. If we refer to Section \ref{subsec:hierarchical_exp2_datasets}, we see that the average number of sentences per document in the \emph{Multi-domain Customer Review} and \emph{Reuters} dataset is 3.78 and 7.40 respectively, which is much lower than that of the other datasets and this leads to a situation in which HBM cannot effectively extract the sentence structure information.

%
%

We can conclude that compared to several baseline methods, the proposed Hierarchical BERT Model gives superior performance across the datasets containing long documents, which is evidenced by the higher AUC scores. This clearly demonstrates the effectiveness of the proposed model when limited labelled data is available, but also highlights that the approach only works well in domains with multiple sentences per document.

\section{Extracting Salient Sentences Using a HBM Model}\label{sec:hierarchical_exp3}

As mentioned previously, one advantage of the HBM approach is that the attention scores assigned to sentences in a document can be used as an indication of the importance, or saliency, of those sentences in determining the class the document belongs to. The value of this could be that, the low attention sentences can be ignored and the high attention sentences can be selected to highlight to the user to reduce the effort required to annotate a document. In other words, highlighting the salient sentences in a document can be used to facilitate an explanation of the classifications that the model produces. In this section, we will introduce how the HBM infer salient sentences.

\subsection{How to Infer Salient Sentences}\label{sec:hierarchical_exp3_infer}

As described in Equation \ref{eq:attention_score}, $ \text{Softmax}(Q_{i}\times K_{i}^{\top}/\sqrt{d_{e}})$ is a matrix of attention weights used to compute the weighted average of the self-attention heads in the BertAtt layer of HBM. In this matrix, entry $e^{aj}$ (at the $a^{th}$ row and $j^{th}$ column) is the attention weight that the $a^{th}$ sentence pays to the $j^{th}$ sentence. In other words, the $j^{th}$ column represents all the attention weights the $j^{th}$ sentence received from other sentences and we adopt the sum of all entries in the $j^{th}$ column as the saliency score of the $j^{th}$ sentence. Therefore, we can compute the saliency score of each sentence in a document by summing up all entries in the corresponding column of the attention weights matrix.

After obtaining the saliency score of all sentences in a document, we select the salient sentences based on the ratio of the sentence saliency score to the highest saliency score. We regard all sentences with a ratio bigger than 0.9 as salient sentences in a document and disregard other sentences. We believe that, besides the sentence with the highest saliency score, those with a saliency score close to the highest saliency score should also contribute intensively to the document representation and should be highlighted. In the next subsection, we will show the effectiveness of highlighting salient sentences through a user study.

\section{User Study}\label{sec:hierarchical_exp4}

It is interesting to understand to what extent we can take advantage of important sentences identified by the HBM to help the human with the annotation task. Therefore, the objective of this study is to investigate whether or not highlighting the important sentences in a text makes it easier for labelers to provide these category labels, thus facilitating labelling of large numbers of texts in a short time. This addresses the second research question {\fontseries{b}\selectfont RQ2}.

\begin{figure*}[!htb]
\centerline{\includegraphics[width=.8\textwidth]{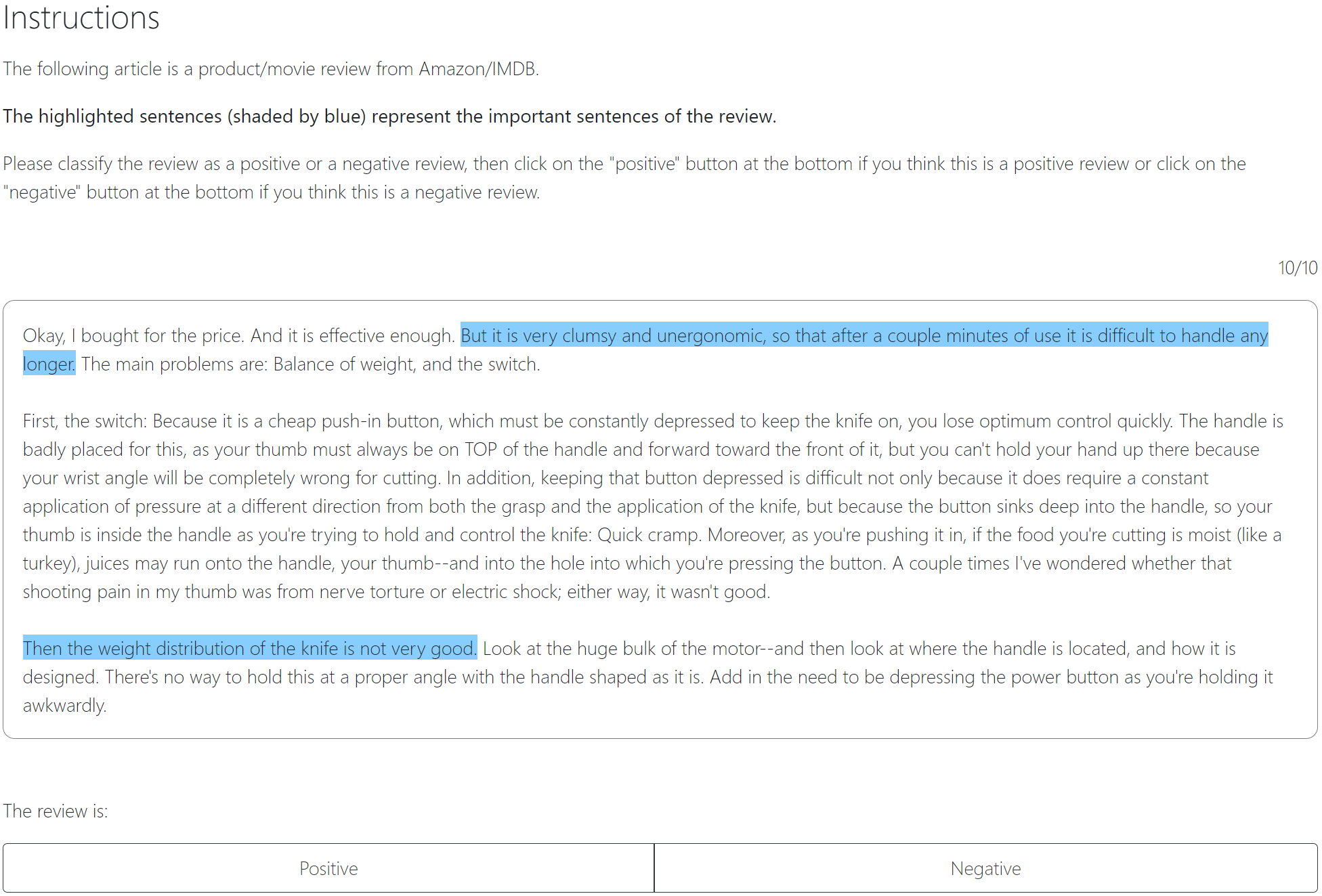}}
 \caption{A screenshot of one page of the user study website showing a document with highlighting.}\label{fig:hierarchical_user_study_sample1}
\end{figure*}

\begin{figure*}[!htb]
\centerline{\includegraphics[width=.8\textwidth]{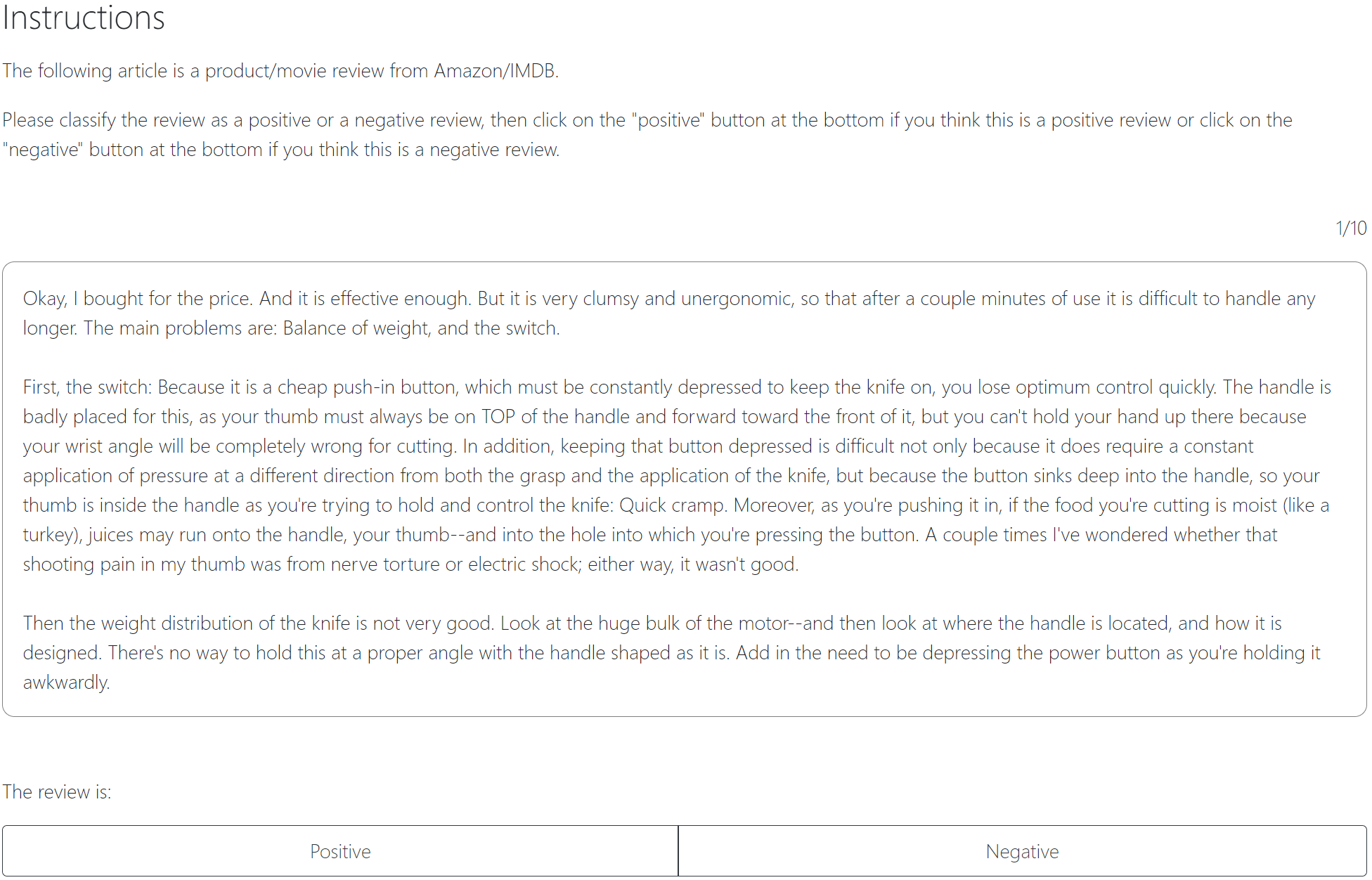}}
 \caption{A screenshot of one page of the user study website showing a document with no highlighting.}\label{fig:hierarchical_user_study_sample2}
\end{figure*}

In this user study, we use the ten documents with highlighted sentences inferred by HBM from the \emph{Movie Review} and \emph{Multi-domain Customer Review datasets} and another ten documents with the same text shown without highlighting.\footnote{The user study platform information and all these ten documents with highlighted sentences are available at https://github.com/GeorgeLuImmortal/Hierarchical-BERT-Model-with-Limited-Labelled-Data} Figures \ref{fig:hierarchical_user_study_sample1} and \ref{fig:hierarchical_user_study_sample2} shows a pair of examples of one of these. 

\vspace{-5mm}

\vspace{-5mm}

\subsection{Experiment Setup}



The user study is run on an online platform and data collected is stored anonymously. We recruited participants by circulating an advertisement among social media such as Twitter and Facebook, and internel email lists of two Irish research centres.\footnote{The Insight Centre for Data Analytics (https://www.insight-centre.org/) and ADAPT Centre (https://www.adaptcentre.ie/).} In this study, all participants were required to be fluent English speakers. Participants are asked to state the categories to which a series of texts (i.e. all ten documents with highlighting or all ten documents without highlighting) belong. They are shown a text and asked to indicate which category, from a list of options (i.e. positive sentiment or negative sentiment) provided, best describes it. Ground truth categories for the texts used in the study exist, and these were used to measure the accuracy of the category labels provided by participants. The time taken for participants to select a category for each document were also collected to measure the annotation efficiency.

A between-subject design was used for the experiment. Half of the participants were shown texts in which important sentences have been highlighted (i.e. highlighting condition group) to make it easier for them to come to a judgement on the category that best describes the text. Half of the participants were shown texts without highlighting (no-highlighting condition group). All documents were presented to the participants in a random sequence. The accuracy of labels provided and the time taken to provide them were compared across the two groups to measure the impact of providing highlighting.

\subsection{Results and Analysis}

We had 75 participants in total: 38 participants in the highlighting condition group and 37 participants in the no-highlighting condition group. To mitigate against participants who didn't pay careful attention to the tasks in the study, we removed the records of participants who achieved the accuracy less than 0.6 or spent less than 90 seconds (i.e. a minute and a half) in total on all ten questions. Similarly, to mitigate against participants who left the study for a long period of time, we also removed data from participants who spent more than 420 seconds (i.e. 7 minutes) on providing the answer for a single document. After this cleaning, we finally have 57 valid participants, among these participants 32 are in highlighting condition group and 25 are in no-highlighting condition group.

\begin{figure*}[ht]
\centerline{\includegraphics[width=.7\textwidth]{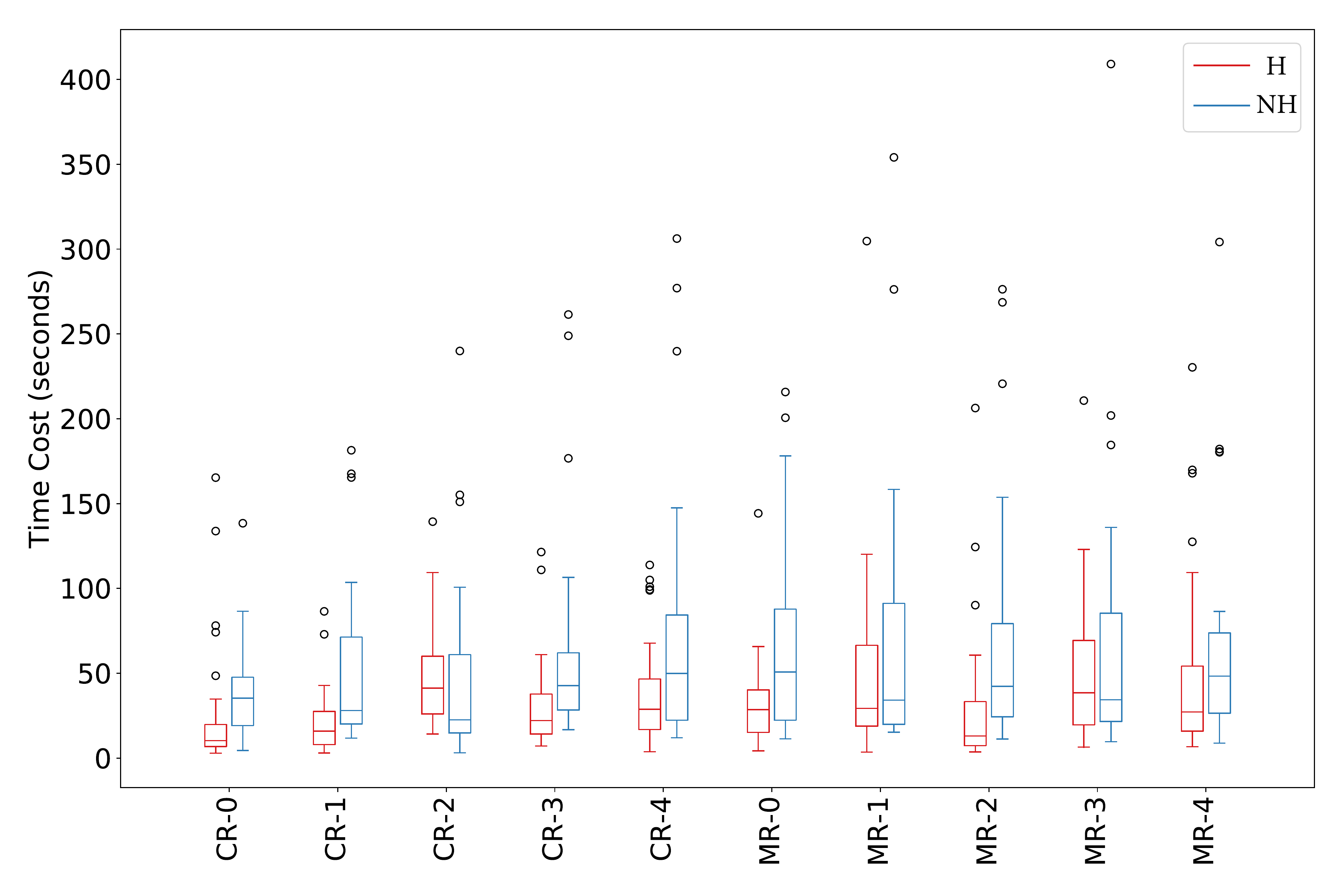}}
 \caption{A box plot of the time cost per user on each document. The vertical axis is the time cost per document (unit second); the horizontal axis represents the documents.}\label{fig:hierarchical_boxplot}
\end{figure*}

Figure \ref{fig:hierarchical_boxplot} shows a box plot of the time spent per user in providing category labels for each document, where red boxes denote the performance of the highlighting condition group, and blue boxes denote the performance of the no-highlighting condition group. It is obvious that except for documents CR-2 and MR-3, the time cost in the highlighting condition group is much less than that in the no-highlighting condition group.

We also analyse the results from the user perspective, the average total time cost in labelling all ten documents per user in the highlighting condition group is 375.50 seconds, and in the no-highlighting condition group is 645.54 seconds. Following the advice in \cite{marusteri2010comparing} we analyse these results using a Mann Whitney U test in which the $p$-value is 0.01939, which indicates that we can reject the null hypothesis at the 0.05 confidence level and consider a significant difference between the time cost of the two groups. The average accuracy achieved by participants in the highlighting condition group is 0.9575, while the average accuracy achieved by participants in the no-highlighting condition group is 0.9448. The $p$-value of a Mann-Whitney U test on average accuracy is 0.3187, which indicates no significant difference between the average accuracy of these two groups exists.

Based on the previous analysis, we conclude that highlighting important sentences identified by the HBM can significantly improve the human's annotation efficiency while still maintaining a high annotation quality. 

\vspace{-3mm}

\section{Conclusions}\label{sec:hierarchical_summary}

\vspace{-3mm}

In this work, we have proposed a novel approach, the Hierarchical BERT Model (HBM), which has been demonstrated to perform well with limited labelled data especially when documents are long. Several experiments have been carried out to demonstrate the superior performance of the HBM as compared to other state-of-the-art methods under the assumption that training sets are small. We also demonstrated that using a HBM faciliates selecting salient sentences from a document and showed through a user study that highlighting these sentences leads to more efficient labelling.

%
%
\bibliographystyle{splncs04}
\bibliography{DS_2021_HBM}

\clearpage
\end{document}